# Image Resolution and Contrast Enhancement of Satellite Geographical Images with Removal of Noise using Wavelet Transforms

Prajakta P. Khairnar*[1], Prof. C. A. Manjare*[2]

[1] M.E. (Electronics (Digital Systems) second year), PG student, Jaywantrao Sawant College of Engg, Hadapsar, Pune, India

[2] Assistant Professor, Department of E & TC, Jaywantrao Sawant College of Engg, Hadapsar, Pune, India

[1] khairnarprajakta@yahoo.co.in

**Abstract--**In this paper the technique for resolution and contrast enhancement of satellite geographical images based on discrete wavelet transform (DWT), stationary wavelet transform (SWT) and singular value decomposition (SVD) has been proposed. In this, the noise is added in the input low resolution and low contrast image. The median filter is used remove noise from the input image. This low resolution, low contrast image without noise is decomposed into four sub-bands by using DWT and SWT. The resolution enhancement technique is based on the interpolation of high frequency components obtained by DWT and input image. SWT is used to enhance input image. DWT is used to decompose an image into four frequency sub bands and these four sub-bands are interpolated using bicubic interpolation technique. All these sub-bands are reconstructed as high resolution image by using inverse DWT (IDWT). To increase the contrast the proposed technique uses DWT and SVD. GHE is used to equalize an image. The equalized image is decomposed into four sub-bands using DWT and new LL sub-band is reconstructed using SVD. All sub-bands are reconstructed using IDWT to generate high resolution and contrast image over conventional techniques. The experimental result shows superiority of the proposed technique over conventional techniques.

## I. INTRODUCTION

Resolution and contrast are the most important aspect of an image. One of the commonly used techniques for the resolution enhancement is interpolation [4][13]. Some of the well-known techniques are bicubic interpolation, bilinear interpolation and nearest neighbor interpolation. The main disadvantage in using interpolation is the loss of high frequency (HF) components which is due to the smoothing caused by interpolation. To avoid this loss, we use a tool called wavelet transform. We use wavelet transforms because of their inherent property of redundant and shift invariant. Another wavelet used in image processing is Stationary Wavelet Transform (SWT)[8]. Discrete Wavelet Transform (DWT) and Stationary Wavelet Transform (SWT) are used to decompose the low resolution input image into frequency component or sub-bands namely LL, LH, HL and HH bands. Here, LL sub-band consists of illumination information and remaining sub-bands constitutes the information of edges. SWT is same as DWT but it does not use down sampling, hence the sub-bands of SWT and the input image are having same size.

For contrast enhancement we have some basic operations like general histogram equalization (GHE), local histogram equalization (LHE) and brightness preserving dynamic histogram equalization (BPDHE)[7]. If contrast of image is highly concentrated on a specific range, the information may be lost in those areas which are uniformly concentrated. To overcome this issue above histogram equalization techniques are used. These techniques help to optimize the contrast of an image in order to represent all information in the input image. GHE[13] is widely used and simple contrast enhancement technique in which the output histogram is uniformly distributed. One of the disadvantages of GHE is that the information on the histogram input image will be lost. To overcome this Singular Value Decomposition (SVD) is introduced. We use SVD [5] with DWT algorithm.

The two main advantages of using SVD in proposed technique are: First, the singular value matrix obtained by SVD contains illumination information. Therefore, changing the singular values will directly affect the illumination of the image. Hence, the other information in the image will not be changed. Second, by applying the illumination enhancement in LL sub-band will protect the edge information in other sub-bands (i.e. LH, HL, and HH).

II. PROPOSED IMAGE ENHANCEMENT TECHNIQUE

Basically there are two steps involved in this enhancement technique. The first step is the resolution enhancement using DWT and SWT and the second step is the contrast enhancement using SVD.

*A. Resolution enhancement*

In the proposed method, the noise is added in the input low resolution and low contrast image. The median filter is used remove noise from the input image[14]. The proposed enhancement process is based on the interpolation of HF sub-band images obtained by DWT and input image. In image resolution enhancement by using interpolation, the main loss is on its high frequency components (i.e. edges), which is due to the smoothing caused by interpolation. In order to increase the quality of the image, preserving the edges is essential. In this work, DWT is used in order to preserve the high frequency components of the image. The redundancy and shift invariance of the DWT mean that DWT coefficients are inherently interpolable[2]. DWT is used to decompose an input image into four different sub-band images[11]. Three high frequency sub-bands (LH, HL, and HH) contain the high frequency components of the input image. In the proposed technique, bicubic interpolation with enlargement factor of 2 is applied to high frequency sub-band images. Down sampling in each of the DWT sub-bands causes information loss in the respective sub-bands. To minimize this loss SWT is used.

The interpolated high frequency sub-bands and the SWT high frequency sub-bands have the same size which means they can be added with each other in order to correct the estimated coefficient. Also it is known that in the wavelet domain, the low resolution image is obtained by low pass filtering of the high resolution image[11]. In other words, low frequency sub-band is the low resolution of the original image. Therefore, instead of using low frequency sub-band, which contains less information than the original high resolution image, we are using the input image for the interpolation of low frequency sub-band image. i.e. input image is also interpolated separately. By combining corrected HF sub-bands and interpolated input image through IDWT, high resolution output is achieved. The output image will contain sharper edges than the interpolated image obtained by interpolation of the input image directly. This is due to the fact that, the interpolation of isolated high frequency components in high frequency sub-bands and using the corrections obtained by adding high frequency sub-bands of SWT of the input image, will preserve more high frequency components after the interpolation than interpolating input image directly.

By applying the 1-D discrete wavelet transform (DWT) along the rows of the image first, and then along the columns to produce 2-D decomposition of image. DWT produce four sub-bands low-low (LL), low-high (LH), high-low (HL) and high-high (HH). By using these four sub-bands we can regenerate original image. Theoretically, a filter bank shown in Fig. 1 should work on the image in order to generate different sub-band frequency images.

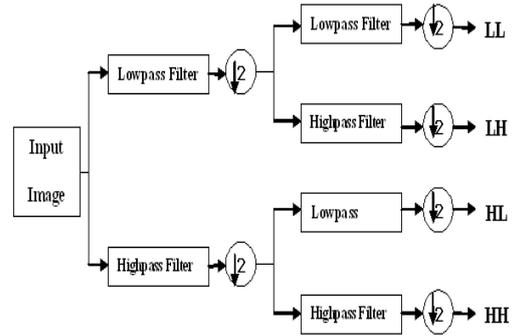

Fig.1. Block diagram of DWT Filter Banks of level 1

*B. Contrast enhancement*

The output of the resolution enhancement module will be taken as the input to this module. For contrast enhancement the main tools that we use here are SVD and DWT. Singular value matrix obtained by SVD contains the illumination information. So we have to change this matrix to change the contrast of the image. Any changes made to this matrix will not affect the other attributes of the image. The DWT is used to divide the image into sub-bands. We know that the edges are concentrated on LH, HL and HH sub-bands. Hence, even if we separate the HF components and apply some transformations on the LF will not cause any damage to the edge components. Hence after reconstruction the image looks too sharper. The key steps involved in this process are as follows:

First we will apply DWT on the input image and in parallel we improve its contrast using GHE and find its DWT. Now calculate the hanger (U), aligner (V) and singular value matrix (SVM) for the LL sub-bands obtained above. Singular value decomposition of an image which can be interpreted as a matrix is written as follows-

$$A = U_A \Sigma_A V_A^T \qquad (1)$$

where, $U_A$ and $V_A$ are orthogonal square matrices known as Hanger and aligner, respectively and the ΣA matrix contain the sorted singular values on its main diagonal. The idea of using SVD[5] for image equalization comes from this fact that ΣA contains the intensity information of a given image. The method uses the ratio of largest singular value of the generated normalized matrix, with zero mean and unity variance of, over a normalized image which can be calculated according to

$$\xi = \frac{Max(\Sigma_N (\mu=0, var=1))}{Max(\Sigma_A)} \qquad (2)$$

where, ΣN is the singular value matrix and the synthetic intensity matrix at zero mean and unity variance. This coefficient can be used to regenerate an equalized image using "(1)".

$$\Sigma EqualizedA = U_A(\xi \Sigma_A)V_A^T \qquad (3)$$

Here, we take the help of DWT to decompose this image into different sub-bands. The resultant can be obtained by combining the sub-band images using IDWT.

On finding the maximum element in both the SVMs and their ratio (ξ), now calculate the new STM and estimate the new LL sub-band.

$$LL_{A(new)} = U_{LLA}\Sigma_{LLA}V_A^T \qquad (4)$$

The estimated LL sub-band and the HF components of actual input image are now used to re-produce the contrast enhanced image. From the close observation we can see that the HF components are not disturbed. We manipulated the illumination information alone. Hence we can be sure that there is no harm to edge components.

### III. BLOCK DIAGRAM

Fig.2. shows the block diagram of proposed system. In this figure, the noise is added in the input low resolution and low contrast image. The median filter is used remove noise from the input image. This low resolution, low contrast image without noise is decomposed into four sub-bands by using DWT and SWT. The HF sub-bands of DWT are interpolated by using bicubic interpolation to get same size as that of SWT sub-bands. HF components of both DWT and SWT are added for correction. Then these corrected HF components are given to IDWT with input image to get high resolution low contrast image. This image is equalized by using GHE and then DWT is applied to get four sub-bands. SVD is applied to reconstruct new LL sub-band. All sub-bands are reconstructed using IDWT to generate high resolution and high contrast image.

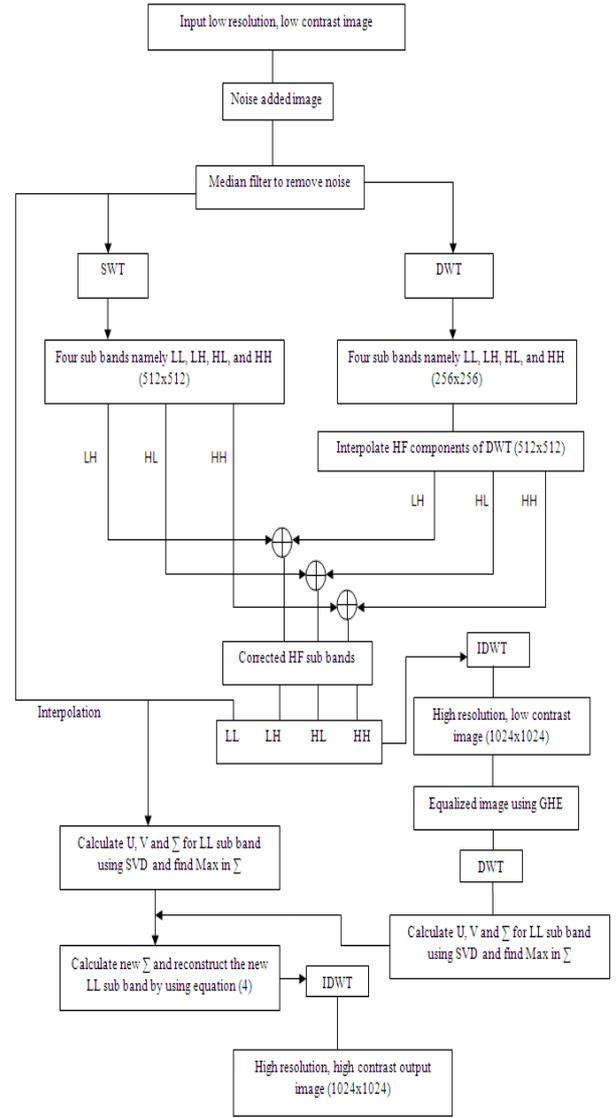

Fig.2. Block diagram of proposed system

### IV. RESULT

In the proposed method, we are using standard image like Lena for presenting the output. Fig 3 shows input and output for Lena image. Fig.4 shows the typical application of the proposed technique with a) Input image, b) Noise added image, c) Noise filtered image, d) Resolution enhanced image, and e) Resolution and contrast enhanced output for the input satellite image.

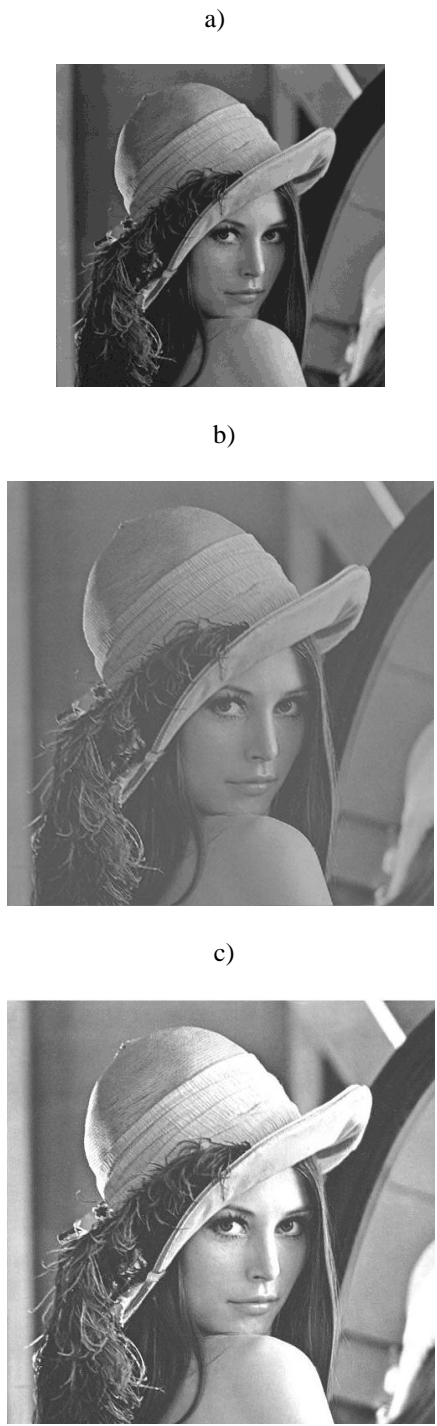

Fig.3. a) Input Image, b) Resolution Enhanced Image, and c) Resolution and Contrast Enhanced Output Image for standard Lena Image.

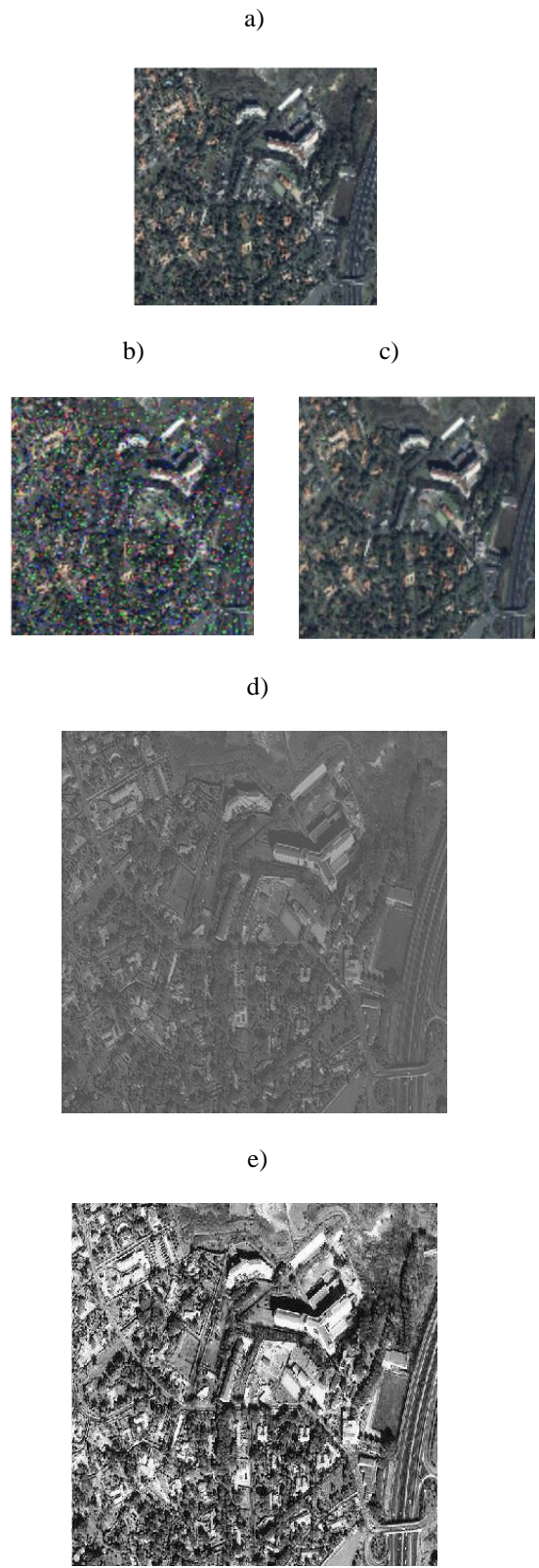

Fig.4. a) Input Image, b) Noise added image, c) Noise filtered image, d) Resolution Enhanced Image, and e) Resolution and Contrast Enhanced Output Image for satellite Input Image

Also, not only observing the input and output images, there is a significant improvement in resolution and contrast to measure the quality of image by calculating the peak signal to noise ratio (PSNR), RMSE, Entropy. Peak signal-to-noise ratio (PSNR) and root mean square error (RMSE) have been implemented in order to obtain some quantitative results for comparison. PSNR can be obtained by using the following formula:

$$PSNR = 10 log_{10}(R^2/MSE) \qquad (5)$$

where, R is the maximum fluctuation in the input image (255 in here as the images are represented by 8 bit, i.e., 8- bit gray scale representation have been used radiometric resolution is 8 bit).

$$MSE = \frac{\sum_{i,j}\left(I_{in}(i,j) - I_{org}(i,j)\right)^2}{M \times N} \qquad (6)$$

where M and N are the size of the images. When the two images are identical, the MSE will be zero. Clearly, RMSE is the square root of MSE, hence it is given by:

$$RMSE = \sqrt{MSE} \qquad (7)$$

Image entropy is a quantity which is used to describe the `business' of an image, i.e. the amount of information which must be coded for by a compression algorithm. Image entropy is calculated with the formula:

$$ENTROPY = -\sum Pi log_2 Pi \qquad (8)$$

In the above expression, Pi is the probability that the difference between two adjacent pixels is equal to i, and $log_2$ is the base 2 logarithms.

TABLE I

PSNR value for Lena image and satellite image

| Sr. No. | Image | PSNR value in dB |
|---|---|---|
| 1. | Lena Image (Fig. 3) | 36.49 |
| 2. | Satellite image (Fig. 4) | 36.19 |

V. CONCLUSIONS

This work, proposed a new image resolution enhancement technique based on the interpolation of the high frequency sub-bands obtained by DWT and correcting HF sub-band using SWT with the input image. Contrast enhancement is based on the reconstruction of LL sub-band obtained by DWT, using singular value matrix which gives illumination contents. This application is more prominent in satellite image processing for disaster management. We can uncover the hidden details from the given images. Also, the PSNR value is more than the conventional methods.